\documentclass{article}
\usepackage{spconf,amsmath,graphicx,hyperref}

\usepackage{algorithm}
\usepackage{algorithmic}
\usepackage{amssymb}
\usepackage{tikz}
\usepackage{booktabs}
\usepackage{siunitx}
\usepackage{subcaption}
\usepackage[percent]{overpic}
\title{Shared-Weights Extender and Gradient Voting for Neural Network Expansion}
\name{Nikolas Chatzis$^{1,2,3}$, Ioannis Kordonis$^{3}$, Manos Theodosis$^{4}$, Petros Maragos$^{1,2,3}$
}
\address{
\small $^1$Robotics Institute, Athena Research Center, 15125 Maroussi, Greece\\
\small $^2$HERON - Hellenic Robotics Center of Excellence, Athens, Greece\\
\small $^3$School of ECE, National Technical University of Athens, Greece\\
\small $^4$School of Engineering and Applied Sciences Harvard University Cambridge, MA 02138\\
{\tt\footnotesize n.chatzis@athenarc.gr, kordonis@central.ntua.gr,  etheodosis@seas.harvard.edu, maragos@cs.ntua.gr}
}

\begin{document}
\ninept
\maketitle
\begin{abstract}
Expanding neural networks during training is a promising
way to augment capacity without retraining larger models
from scratch. However, newly added neurons often fail to
adjust to a trained network and become inactive, providing no contribution to capacity growth. We propose the
Shared-Weights Extender (SWE), a novel method explicitly designed to prevent inactivity of new neurons by coupling them with existing ones for smooth integration. In parallel, we introduce the
Steepest Voting Distributor (SVoD), a gradient-based method
for allocating neurons across layers during deep network
expansion. Our extensive benchmarking on four datasets shows that our method can effectively
suppress neuron inactivity and achieve better performance
compared to other expanding methods and baselines.

\end{abstract}

\section{Introduction}
\label{sec:intro}

When designing machine learning models, a central consideration is ensuring that the architecture has sufficient capacity for the task at hand. By capacity, we refer to the ability of a model to represent complex functions and learn from data effectively. Capacity is determined primarily by two aspects of an architecture: its structural design (e.g., layer types and connectivity) and its size (i.e., the number of parameters). A large body of research has addressed the first aspect under the umbrella of Neural Architecture Search (NAS) and its many variants, where reinforcement learning, evolutionary algorithms, or differentiable optimization are used to explore alternative designs (\cite{zoph2017neural, pham2018enas,liu2018darts,10.1007/978-3-030-58555-6_28}).  

In contrast, deciding the appropriate size of a model is often done empirically. A common strategy is to begin with a large, overparameterized network and then compress it to remove redundancy \cite{Han2015LearningBW}. Here, we explore the opposite direction: what if a trained network is found to lack capacity? Can we adjust its size dynamically during training? A complementary line of research suggests that this is possible, and that capacity can be adjusted during training by dynamically expanding a network in width or depth. In fact, expanding during training may even yield better results than training a larger model from scratch in some cases \cite{NEURIPS2020_fdbe012e,evci2022gradmax}. 
Network expansion has also been investigated in continual learning, where additional capacity is introduced to accommodate new tasks while mitigating catastrophic forgetting (\cite{Rusu2016ProgressiveNN,yoon2018lifelong,NEURIPS2020_fdbe012e}).

\subsection{Related Work}
\label{subec:format}

Several works have followed a natural path to guide the extension of a network: The gradient of the loss with respect to newly inserted parameters can be used to quantify how effective an insertion is. 
In \cite{Liu2019SplittingSD}, the authors introduce the Splitting Steepest Descent method, which uses second-order gradient information to decide which neurons to split and how to split them.
\cite{wang2020energyawareneuralarchitectureoptimization} extends this approach into a more efficient scheme.
The Firefly Algorithm \cite{NEURIPS2020_fdbe012e} generalizes this idea by selecting among a set of architecture modifications, such as splitting existing neurons or adding new ones, based on their estimated gradient impact.
Similarly, GradMax in \cite{evci2022gradmax} adds neurons in a way that preserves the current network function and maximizes the gradient magnitude of the loss.
Self Expanding Neural Networks (SENN) in \cite{mitchell2024selfexpandingneuralnetworks} use a formal metric, the natural expansion score, which quantifies how much a candidate expansion is expected to reduce the loss. 

Beyond optimization-driven approaches, several methods rely on heuristic rules to guide architectural changes during training. 
When expanding a network, a key observation is that the augmented architecture should preserve the essential information learned by the original model. In this direction, Net2Net \cite{Chen2015Net2NetAL} expands networks by inserting identity-initialized layers and widening layers through neuron replication. In another approach \cite{cimb44020056}, an adaptive training algorithm is proposed for constructing compact, task-specific neural architectures by alternating between adding new hidden layers and pruning weak connections based on neuron activity.
Focusing only in Continual Learning Tasks, in Progressive Neural Networks \cite{Rusu2016ProgressiveNN} for each new task, a fresh set of neurons is added at every layer, while all previously learned parameters are kept frozen.
Similarly, Dynamically Expandable Networks \cite{yoon2018lifelong} start with a compact model and adapt to new tasks through selective retraining and neuron addition, allowing the model to grow only when representation capacity is insufficient.

\subsection{Contribution}
\label{subsec:contribution}

In this paper we propose a method that combines elements from both directions and is tailored to single-task settings, focusing on networks with ReLU activations. A central challenge in this scenario is that newly added neurons often become inactive shortly after insertion. By inactive neurons, we refer to ReLU units whose outputs are identically zero across the dataset, rendering them unable to contribute to learning. While this phenomenon has been studied in networks trained from scratch \cite{deadrelu}, to the best of our knowledge no expansion method explicitly addresses it, even though it is particularly severe for newly inserted neurons, as we demonstrate. Our Shared-Weights Extender (SWE) mitigates this issue by ensuring smooth integration of new neurons through communication with previously trained neurons in the same layer. In parallel, our Steepest Voting Distributor (SVoD) allocates neurons across layers during deep network growth, complementing the insertion procedure. Extensive experiments demonstrate the effectiveness of our method in integrating new neurons and its competitiveness against other expanding approaches. Code for reproducing our experiments is available at: \texttt{https://github.com/kvantonikolas/SWE-SVoD}.

\section{METHOD}
\label{sec:pagestyle}

\subsection{Preliminaries}
\label{subsec:Preliminaries}

We consider a fully connected network $N$ with ReLU activations and $k$ layers, though the underlying ideas are not restricted to this setting. Each layer $l$ contains $n_l$ neurons with weights $w_{ij}^{(l)}$ and biases $b_i^{(l)}$. Training seeks to minimize the empirical loss over a dataset $\mathcal{D} = \{(x_i, y_i)\}_{i=1}^n$.  
Our framework progressively grows the network by alternating between expansion and training. In order to add new neurons, we split the expansion into stages. At the beginning of each stage the new neurons for this stage are inserted. Each layer will be widened by $\{m_l\}_{l=1}^k$ neurons. After insertion, the weights of both the new and existing neurons are adjusted, and then the entire network is trained jointly. This expansion–training cycle is repeated until all stages are completed.

\subsection{Framework}
\label{subsec:Components}
We distinguish the two different operations by introducing two key components: \textbf{Extender} $\mathbf{E}$, which initializes new and adjusts both old and new weights and the \textbf{Distributor} $\mathbf{D}$, which determines the number of neurons at each layer at a certain stage. Our work focuses on the design of these components, which are intended to facilitate effective integration of newly added neurons.
For the Extender we propose the Shared-Weights Extender (SWE), which smoothly incorporates new neurons and addresses the common issue of them becoming inactive early.
For the Distributor, we propose Steepest Voting Distributor (SVoD), inspired from the Firefly method \cite{NEURIPS2020_fdbe012e} that identifies which neurons' insertion reduces loss more effectively. 
Our general framework can be summarized in Algorithm \ref{alg:framework_algo} .



\begin{algorithm}[H]
\caption{Framework for Growing a Neural Network}
\label{alg:framework_algo}
\begin{algorithmic}[1]
\STATE \textbf{Input:} Initial network $N$, dataset $\mathcal{D}$, total neurons to add, number of stages, and helper functions $\mathbf{D,E}$ for distributing and initializing new neurons.
\FOR{each stage}
    \STATE Decide how many neurons to insert at each layer with $\mathbf{D}$.
    \STATE Initialize the new neurons and adjust the existing ones with $\mathbf{E}$ at each layer separately.
    \STATE Jointly train the expanded network $N$.
\ENDFOR
\STATE \textbf{Output:} Final expanded network $N$.
\end{algorithmic}
\end{algorithm}

\subsection{Extender}
\label{subsec:extender}

The intuition behind the Shared-Weights Extender (SWE) is to facilitate the smooth integration of newly added neurons into the network. As shown in Table~\ref{tab:inactivity}, newly inserted neurons often become inactive across the dataset after only a few training epochs, not contributing to the model at all. To address this issue, SWE introduces a short adjustment phase in which new neurons temporarily share parameters with existing ones, allowing them to adapt to previously learned representations before full training resumes.

As defined above, each layer contains $n_l$ neurons with parameters $(\mathbf{w}_i, b_i), \, i \in [n_l]$. 
For clarity, we omit the layer index notation in what follows. 
To integrate a new neuron, we introduce $n_l$ auxiliary parameter pairs $(\mathbf{w}_{c,i}, b_{c,i})$, all initialized at zero. 
The outgoing weights of the new neuron are likewise initialized to zero. 
During this phase, the effective parameters employed by the network for the new neuron $(\mathbf{w}^{\text{eff}}_{\text{new}}, b^{\text{eff}}_{\text{new}})$ and for the existing ones $(\mathbf{w}^{\text{eff}}_{i}, b^{\text{eff}}_{i})$ are defined as follows:

\begin{equation}
\mathbf{w}^{\text{eff}}_{\text{new}} = \mathbf{w}_{\text{new}} + \sum_{i=1}^{n_l} \mathbf{w}_{c,i}, 
\quad 
b^{\text{eff}}_{\text{new}} = b_{\text{new}} + \sum_{i=1}^{n_l} b_{c,i},
\label{eq:new}
\end{equation}

\begin{equation}
\mathbf{w}^{\text{eff}}_{i} = \mathbf{w}_i - \mathbf{w}_{c,i}, 
\quad 
b^{\text{eff}}_{i} = b_i - b_{c,i}.
\label{eq:old}
\end{equation}

\noindent In this formulation, each existing neuron donates a small, learnable portion of its activation to the new neuron through the coupling terms $(\mathbf{w}_{c,i}, b_{c,i})$. Only the coupling parameters are updated during this phase, along with the parameters of the next layer. 

\sloppy{
We experimentally observed that a single forward-backward pass is sufficient to initialize the couplings effectively. After this step, the couplings are merged into the base weights as in Eqs.~\ref{eq:new} and \ref{eq:old}, the auxiliary parameters $(\mathbf{w}_{c,i}, b_{c,i})$ are discarded, and training resumes with the updated parameters as the new starting point, as described in Algorithm~\ref{alg:framework_algo}.We also note that for expanding deep architectures, this procedure is applied layer by layer, so the total cost amounts to one forward pass per layer.
}

\begin{figure}[h]
  \centering
  \resizebox{0.7\linewidth}{!}{%
    \tikzset{every picture/.style={line width=0.75pt}} 

\begin{tikzpicture}[x=0.75pt,y=0.75pt,yscale=-1,xscale=1]

\draw [color={rgb, 255:red, 0; green, 0; blue, 0 }  ,draw opacity=0.3 ]   (296.48,129.44) -- (416.48,78.69) ;
\draw [color={rgb, 255:red, 0; green, 0; blue, 0 }  ,draw opacity=0.3 ]   (296.48,129.44) -- (416.48,122.8) ;
\draw [color={rgb, 255:red, 0; green, 0; blue, 0 }  ,draw opacity=0.3 ]   (296.48,129.44) -- (416.48,177.9) ;
\draw [color={rgb, 255:red, 0; green, 0; blue, 0 }  ,draw opacity=0.3 ]   (296.38,178.62) -- (416.48,78.69) ;
\draw [color={rgb, 255:red, 0; green, 0; blue, 0 }  ,draw opacity=0.3 ]   (296.38,178.62) -- (416.48,122.8) ;
\draw [color={rgb, 255:red, 0; green, 0; blue, 0 }  ,draw opacity=0.3 ]   (416.48,177.9) -- (296.38,178.62) ;
\draw [color={rgb, 255:red, 0; green, 0; blue, 0 }  ,draw opacity=0.3 ]   (456.59,78.69) -- (527.89,159.6) ;
\draw [color={rgb, 255:red, 0; green, 0; blue, 0 }  ,draw opacity=0.3 ]   (456.59,122.8) -- (527.89,159.6) ;
\draw [color={rgb, 255:red, 0; green, 0; blue, 0 }  ,draw opacity=0.3 ]   (456.59,177.9) -- (527.89,159.6) ;
\draw [color={rgb, 255:red, 208; green, 2; blue, 27 }  ,draw opacity=0.34 ]   (456.59,222.11) -- (527.89,159.6) ;
\draw  [color={rgb, 255:red, 255; green, 255; blue, 255 }  ,draw opacity=1 ][line width=3] [line join = round][line cap = round] (321.89,118.65) .. controls (325.11,117.4) and (335.44,111.6) .. (339.22,110.42) .. controls (347.3,107.91) and (349.81,107.47) .. (357.83,104.26) .. controls (361.65,102.73) and (365.91,97.83) .. (371.95,97.83) .. controls (373.55,97.83) and (369.28,99.28) .. (368.1,100.15) .. controls (364.16,103.06) and (353.98,110.61) .. (353.98,106.31) .. controls (353.98,101.83) and (363.41,104.41) .. (366.49,101.95) .. controls (366.88,101.64) and (366.34,100.25) .. (366.17,100.66) .. controls (365.72,101.75) and (365.26,104.56) .. (364.25,103.74) .. controls (363.6,103.22) and (366.26,102.6) .. (365.85,101.95) .. controls (365.48,101.35) and (364.52,103.7) .. (363.93,103.23) .. controls (363.34,102.76) and (366.11,99.52) .. (365.21,101.69) ;
\draw  [color={rgb, 255:red, 255; green, 255; blue, 255 }  ,draw opacity=1 ][line width=3] [line join = round][line cap = round] (366.81,101.17) .. controls (363.63,103.36) and (359.24,107.52) .. (354.3,108.37) .. controls (350.1,109.09) and (345.59,108.42) .. (341.47,109.4) .. controls (337.79,110.27) and (333.23,111.28) .. (331.52,114.02) ;
\draw  [color={rgb, 255:red, 255; green, 255; blue, 255 }  ,draw opacity=1 ][line width=3] [line join = round][line cap = round] (327.67,114.02) .. controls (334.98,129.62) and (328.64,140.66) .. (323.18,156.16) ;
\draw  [color={rgb, 255:red, 255; green, 255; blue, 255 }  ,draw opacity=1 ][fill={rgb, 255:red, 255; green, 255; blue, 255 }  ,fill opacity=1 ] (324.56,57.3) -- (393.2,57.3) -- (393.2,268) -- (324.56,268) -- cycle ;
\draw [color={rgb, 255:red, 208; green, 2; blue, 27 }  ,draw opacity=1 ]   (296.38,178.62) -- (416.48,222.11) ;
\draw [color={rgb, 255:red, 208; green, 2; blue, 27 }  ,draw opacity=1 ]   (296.48,129.44) -- (416.48,222.11) ;
\draw [color={rgb, 255:red, 65; green, 117; blue, 5 }  ,draw opacity=1 ]   (416.48,222.11) .. controls (353.02,169.51) and (349.73,126.68) .. (415.48,79.4) ;
\draw [shift={(416.48,78.69)}, rotate = 144.54] [color={rgb, 255:red, 65; green, 117; blue, 5 }  ,draw opacity=1 ][line width=0.75]    (10.93,-3.29) .. controls (6.95,-1.4) and (3.31,-0.3) .. (0,0) .. controls (3.31,0.3) and (6.95,1.4) .. (10.93,3.29)   ;
\draw [color={rgb, 255:red, 65; green, 117; blue, 5 }  ,draw opacity=1 ]   (416.48,222.11) .. controls (377.1,177.7) and (379.15,151.77) .. (415.37,123.65) ;
\draw [shift={(416.48,122.8)}, rotate = 142.65] [color={rgb, 255:red, 65; green, 117; blue, 5 }  ,draw opacity=1 ][line width=0.75]    (10.93,-3.29) .. controls (6.95,-1.4) and (3.31,-0.3) .. (0,0) .. controls (3.31,0.3) and (6.95,1.4) .. (10.93,3.29)   ;
\draw [color={rgb, 255:red, 65; green, 117; blue, 5 }  ,draw opacity=1 ]   (416.48,222.11) .. controls (405.17,203.87) and (407.93,200.75) .. (415.86,179.57) ;
\draw [shift={(416.48,177.9)}, rotate = 110.31] [color={rgb, 255:red, 65; green, 117; blue, 5 }  ,draw opacity=1 ][line width=0.75]    (10.93,-3.29) .. controls (6.95,-1.4) and (3.31,-0.3) .. (0,0) .. controls (3.31,0.3) and (6.95,1.4) .. (10.93,3.29)   ;
\draw  [color={rgb, 255:red, 155; green, 155; blue, 155 }  ,draw opacity=1 ] (416.48,78.69) .. controls (416.48,67.62) and (425.46,58.64) .. (436.53,58.64) .. controls (447.61,58.64) and (456.59,67.62) .. (456.59,78.69) .. controls (456.59,89.77) and (447.61,98.74) .. (436.53,98.74) .. controls (425.46,98.74) and (416.48,89.77) .. (416.48,78.69) -- cycle ;
\draw  [color={rgb, 255:red, 155; green, 155; blue, 155 }  ,draw opacity=1 ] (416.48,122.8) .. controls (416.48,111.72) and (425.46,102.74) .. (436.53,102.74) .. controls (447.61,102.74) and (456.59,111.72) .. (456.59,122.8) .. controls (456.59,133.87) and (447.61,142.85) .. (436.53,142.85) .. controls (425.46,142.85) and (416.48,133.87) .. (416.48,122.8) -- cycle ;
\draw  [color={rgb, 255:red, 155; green, 155; blue, 155 }  ,draw opacity=1 ] (416.48,177.9) .. controls (416.48,166.83) and (425.46,157.85) .. (436.53,157.85) .. controls (447.61,157.85) and (456.59,166.83) .. (456.59,177.9) .. controls (456.59,188.98) and (447.61,197.96) .. (436.53,197.96) .. controls (425.46,197.96) and (416.48,188.98) .. (416.48,177.9) -- cycle ;
\draw  [color={rgb, 255:red, 208; green, 2; blue, 27 }  ,draw opacity=1 ] (416.48,222.11) .. controls (416.48,211.04) and (425.46,202.06) .. (436.53,202.06) .. controls (447.61,202.06) and (456.59,211.04) .. (456.59,222.11) .. controls (456.59,233.19) and (447.61,242.17) .. (436.53,242.17) .. controls (425.46,242.17) and (416.48,233.19) .. (416.48,222.11) -- cycle ;
\draw  [color={rgb, 255:red, 155; green, 155; blue, 155 }  ,draw opacity=1 ] (527.89,159.6) .. controls (527.89,148.52) and (536.87,139.55) .. (547.95,139.55) .. controls (559.02,139.55) and (568,148.52) .. (568,159.6) .. controls (568,170.67) and (559.02,179.65) .. (547.95,179.65) .. controls (536.87,179.65) and (527.89,170.67) .. (527.89,159.6) -- cycle ;
\draw  [color={rgb, 255:red, 155; green, 155; blue, 155 }  ,draw opacity=0.35 ] (464.5,190.85) .. controls (469.17,190.83) and (471.49,188.49) .. (471.47,183.82) -- (471.15,95.64) .. controls (471.12,88.97) and (473.44,85.63) .. (478.11,85.62) .. controls (473.44,85.63) and (471.1,82.31) .. (471.08,75.64)(471.09,78.64) -- (471.04,64.32) .. controls (471.02,59.65) and (468.68,57.33) .. (464.01,57.35) ;

\draw (436.21,144.37) node [anchor=north west][inner sep=0.75pt]  [color={rgb, 255:red, 0; green, 0; blue, 0 }  ,opacity=0.3 ,rotate=-89.63] [align=left] {...\\\\};
\draw (421.55,219.75) node [anchor=north west][inner sep=0.75pt]  [color={rgb, 255:red, 208; green, 2; blue, 27 }  ,opacity=1 ] [align=left] {$\displaystyle new$};
\draw (312.55,211.75) node [anchor=north west][inner sep=0.75pt]  [color={rgb, 255:red, 208; green, 2; blue, 27 }  ,opacity=1 ] [align=left] {$\displaystyle (\mathbf{w}_{\text{new}} ,b_{\text{new}})$};
\draw (480.8,76.33) node [anchor=north west][inner sep=0.75pt]  [color={rgb, 255:red, 0; green, 0; blue, 0 }  ,opacity=0.32 ] [align=left] {old neurons};

\end{tikzpicture} 
  }
  \vspace{-0.5cm}
  \caption{\textbf{Shared Weights Extender}. 
  To add a new neuron, $n_l$ auxiliary parameter pairs are introduced (depicted as green arrows), allowing the new neuron to borrow a learnable portion of the representation from each existing one.}
  \label{fig:swe}
\end{figure}
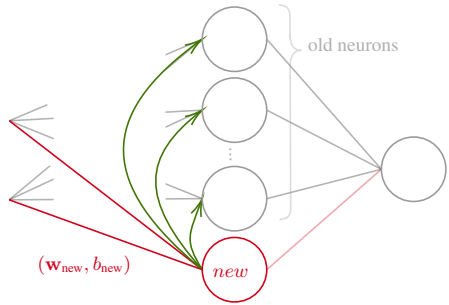

\subsection{Distributor}
\label{subsec:distributor}

Our strategy for allocating new neurons across layers, the Steepest Voting Distributor (SVoD), introduces temporary probe neurons to measure gradient sensitivity. Unlike Firefly, which directly inserts the most effective from the probes it examined, SVoD uses information from all of them to cast votes that identify promising layers, while the actual initialization of new neurons is handled by the Extender.

To quantify the usefulness of a probe neuron, we associate it with a gating variable $z$. Letting $u = \mathbf w^\top \mathbf x + b$ denote its preactivation, the probe’s output is defined as
\begin{equation}
h_{\text{probe}}(z) = \phi\big((1+z)u\big),
\label{eq:probe}
\end{equation}
where $\phi()$ is the activation function. Since $z=0$ leaves the network unchanged, we can analyze the loss as a function of $z$. Expanding around $z=0$ gives
\begin{equation}
\mathcal{L}(z) \approx \mathcal{L}(0) + \frac{\partial \mathcal{L}}{\partial z}\Big|_{z=0}\,z,
\label{eq:loss_taylor}
\end{equation}
A negative value of $\tfrac{\partial \mathcal L}{\partial z}\big|_{z=0}$ indicates that taking a small step in the direction of $(\mathbf{w},b)$ in parameter space would reduce the loss and thus this insertion can be considered effective. 

SVoD identifies promising locations for new neurons by inserting temporary probe neurons into every layer. After a single forward–backward pass, we compute the gradient $\tfrac{\partial \mathcal L}{\partial z}$ for each probe. Only probes with negative gradients are retained, since they indicate that increasing their activation would reduce the loss. The distribution of these negative probes across layers is then used as a voting mechanism: each negative probe casts a vote for the layer it belongs to. Finally, the votes are aggregated and normalized to yield a layer-wise allocation of new neurons. 

\section{EXPERIMENTS}
\label{sec:experiments}

All experiments are conducted on four standard image datasets: MNIST \cite{6296535}, FashionMNIST (FMNIST) \cite{xiao2017fashionmnistnovelimagedataset}, CIFAR-10 \cite{cifar10} and CIFAR-100 \cite{Krizhevsky2009LearningML}. Each experiment is repeated five times, and we report mean and standard deviation unless stated otherwise. We train all networks using the Adam optimizer \cite{adam_optimizer} with a learning rate of \textrm{1e-3}, without additional hyperparameter tuning. Best-performing methods are indicated in bold in the following tables.

\subsection{Results on Neuron Inactivity}
\label{sec:neuroninactivity}

Before diving into demonstrating the performance of our method, we shift our focus on the significant challenge that emerges when trying to add new neurons to an already trained network: newly inserted
neurons tend to become inactive shortly after insertion. Importantly, if a neuron becomes inactive, it will remain so, because the ReLU’s zero derivative blocks gradient flow.  

\begin{table}[htb!]
\centering
\begin{tabular}{@{}lcccc@{}}
\toprule
Neurons & Random & Frobenius & Firefly & SWE (ours)\\
\midrule
 & \multicolumn{4}{c}{\textbf{MNIST}} \\
\cmidrule(lr){2-5}
$20 \rightarrow 40$ & 81.7 & 90.0 &  6.7 & \textbf{0.0} \\
$40 \rightarrow 80$ & 93.2 & 90.0 & 65.8 & \textbf{0.0} \\
$60 \rightarrow 120$& 92.8 & 93.3 & 83.3 & \textbf{0.0} \\
\cmidrule(lr){2-5}
 & \multicolumn{4}{c}{\textbf{FashionMNIST}} \\
\midrule
$20 \rightarrow 40$ & 43.3 & 40.0 & 23.3 & \textbf{0.0} \\
$40 \rightarrow 80$ & 46.7 & 47.5 & 38.3 & \textbf{0.0} \\
$60 \rightarrow 120$& 52.2 & 55.6 & 42.8 & \textbf{0.0} \\
\bottomrule
\end{tabular}
\caption{Inactive neurons $(\%)$ among newly added, after only 5 training epochs. Random denotes standard initialization, while Frobenius denotes norm-preserving initialization. Only SWE manages to reduce the percentage to 0 in this simple task.}
\label{tab:inactivity}
\end{table}

To illustrate this, we train small single hidden-layer networks with 20, 40, and 60 neurons until convergence. After training, we double the number of hidden units by inserting new neurons. We consider three initialization strategies for the new neurons : (i) SWE (ii) the Firefly method (iii) standard random initialization, and (iv) a normalization scheme that tries to mimic SWE idea in a simpler way. Since SWE is designed to preserve the overall contribution of the layer by balancing the weights of old and new neurons, we test a simplified variant where, immediately after insertion, the weights are rescaled so that the Frobenius norm of the updated weight matrix matches that of the original. The extended networks are then trained for 5 additional epochs, and we measure how many of the newly added neurons become inactive. As shown in Table \ref{tab:inactivity}, only SWE ensures the use of all neurons.

\subsection{Results on Image Reconstruction}
\label{sec:imagereconstruction}

To further illustrate the effectiveness of SWE, we evaluate different expanding strategies in the context of growing a single hidden-layer network for autoencoder-style image reconstruction. Starting from a network with $16$ hidden neurons, we incrementally expand the model $7$ times by adding $30\%$ more neurons at each stage. Early stopping is used to determine when to expand the network. As a baseline, we also train static networks with as many hidden neurons as the final expanded models, trained from scratch. Training is performed with the Mean Squared Error (MSE) loss.
 
\begin{figure}[htb!]
  \centering
  \begin{subfigure}[b]{1.0\linewidth}
    \centering
    \begin{overpic}[width=\linewidth,height=3.5cm]{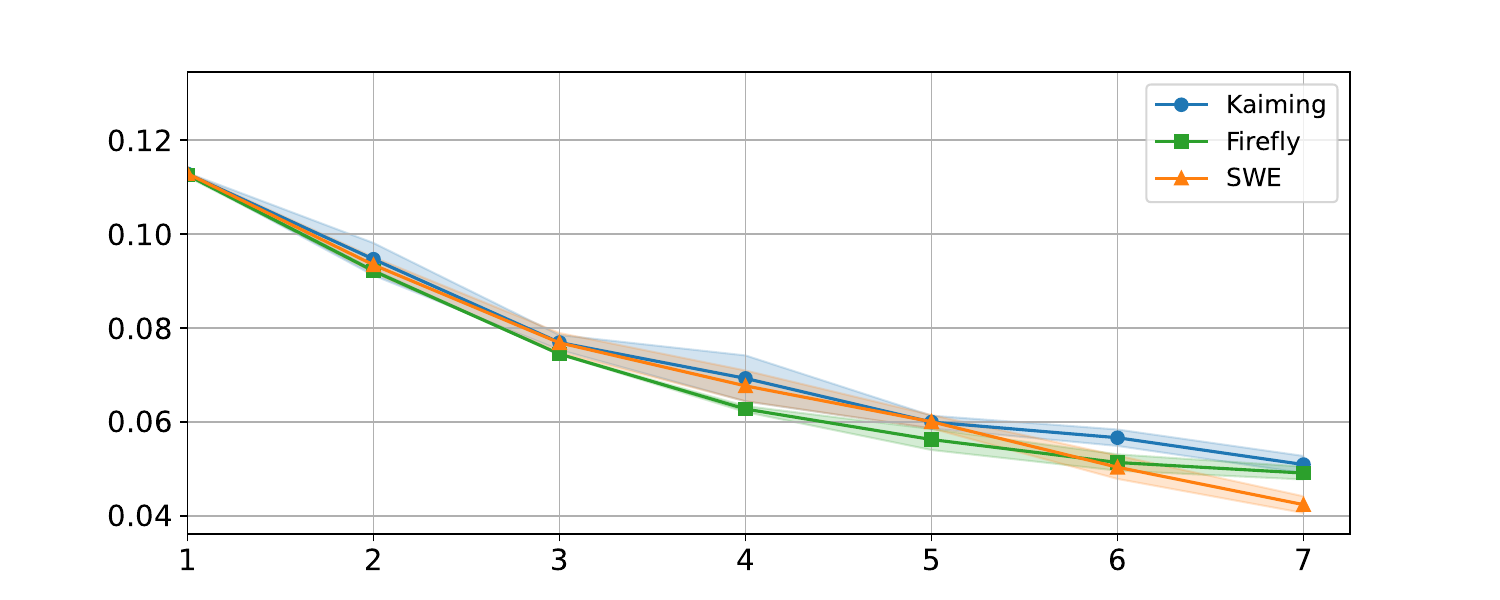}
      \put(3,16){\rotatebox{90}{\text{\scriptsize Loss}}}
      \put(50,0){\makebox(0,0){\text{\scriptsize Stages}}}
    \end{overpic}
    \caption{}
    \label{fig:1hl-rec-mnist}
  \end{subfigure}
  \hfill
  \vspace{-0.4cm}
  \begin{subfigure}[b]{1.0\linewidth}
    \centering
    \begin{overpic}[width=\linewidth,height=3.5cm]{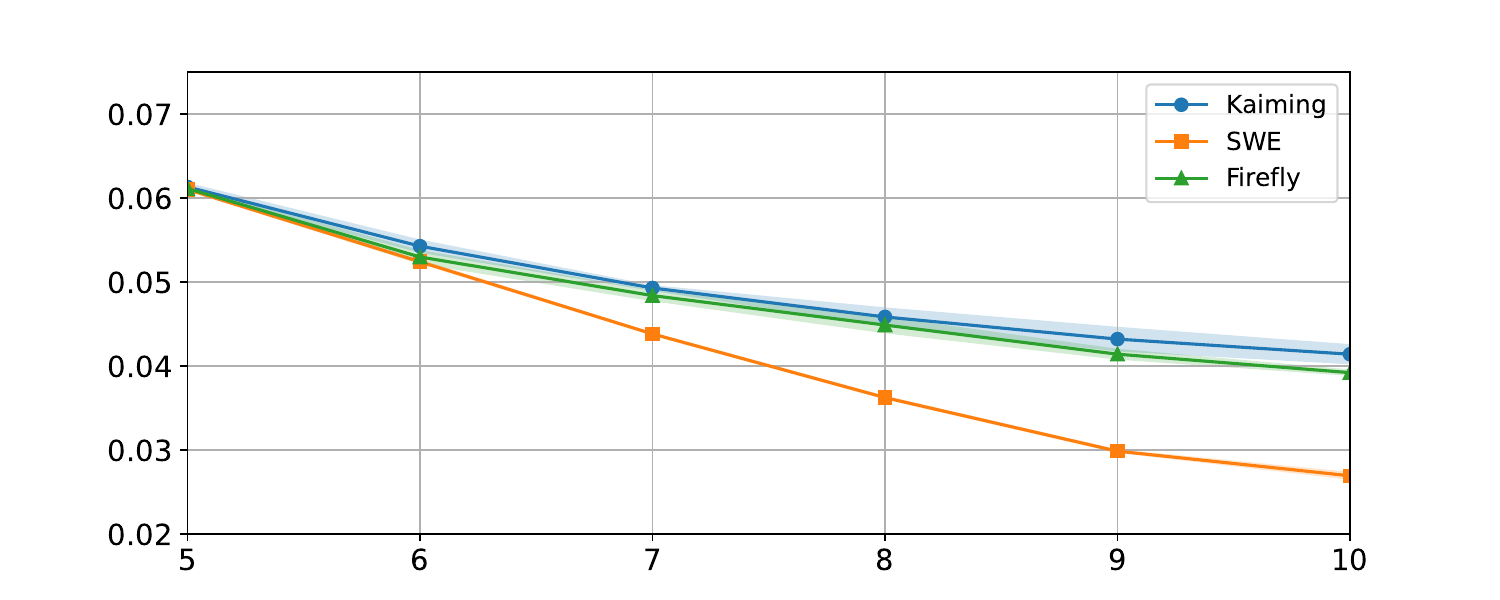}
      \put(3,16){\rotatebox{90}{\text{\scriptsize Loss}}}
      \put(50,0){\makebox(0,0){\text{\scriptsize Stages}}}
    \end{overpic}
    \caption{}
    \label{fig:1hl-rec-fmnist}
  \end{subfigure}

  \caption{Mean test loss during sequential expansions on MNIST (a) and FashionMNIST (b) for the image reconstruction task. 
  SWE achieves lower loss than other methods by a significant margin in the single hidden-layer setting. Shaded regions indicate standard deviation.}
  \label{fig:fmnist-rec}
\end{figure}

In addition to our SWE extender, we evaluate Firefly, where new neurons are sampled from a candidate pool five times larger than the number to be inserted, and the best candidates are selected after training their splitting directions for one epoch. We evaluate also a simple Kaiming initialization \cite{He2015DelvingDI} scheme identical to ours but without the coupling weights mechanism or any smooth transition phase. Final results are summarized in Table~\ref{tab:1hl-recon}. Our proposed SWE consistently achieves lower loss compared to alternative growth strategies, highlighting its effectiveness in integrating newly added neurons. For illustration, Figures~\ref{fig:1hl-rec-mnist} and \ref{fig:1hl-rec-fmnist} depict the progression of reconstruction loss across successive expansion stages. Train set behavior closely follows the training curves and is omitted here for brevity.

\begin{table}[htb!]
\centering
\small
\begin{tabular}{@{}lcccc@{}}
\toprule
 & \multicolumn{4}{c}{\textbf{Method}} \\
\cmidrule(lr){2-5}
\textbf{Dataset} & Baseline & Kaiming & Firefly & SWE  \\
\midrule
MNIST       & $590 \pm 26$ & $509 \pm 18$ & $491 \pm 14$ & $\mathbf{424 \pm 18}$  \\
FMNIST & $ 515 \pm 20$ & $493 \pm 3$  & $484 \pm 7$  & $\mathbf{438 \pm 2}$  \\
\bottomrule
\end{tabular}
\caption{Final test loss $(\times 10^{-4})$ after all stages, for reconstruction with single hidden layer network. SWE outperforms both fixed-size baselines and the Firefly method.}
\label{tab:1hl-recon}
\end{table}

\subsection{Results on Image Classification}
\label{subsec:imageclassification}
\begin{figure}[H]
  \centering
  \begin{subfigure}[b]{1.0\linewidth}
    \centering
    \begin{overpic}[width=\linewidth,height=3.5cm]{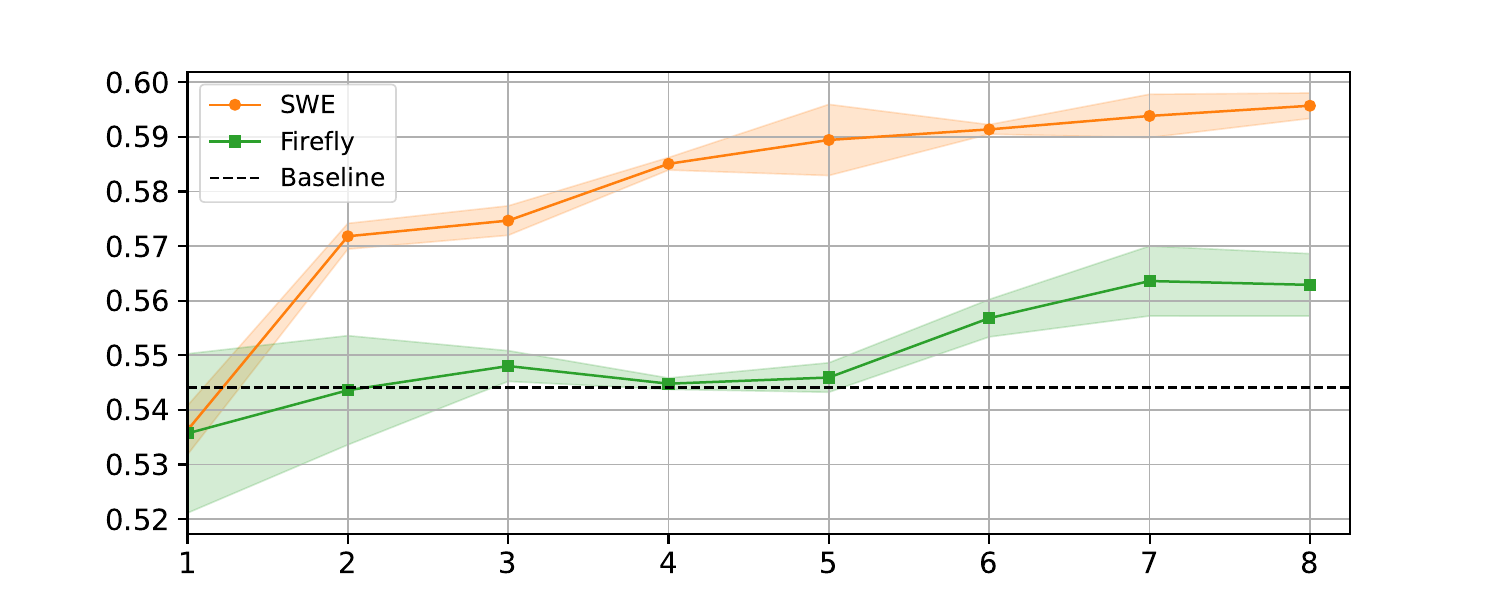}
      \put(3,16){\rotatebox{90}{\text{\scriptsize Accuracy}}}
      \put(50,0){\makebox(0,0){\text{\scriptsize Stages}}}
    \end{overpic}
    \caption{Single hidden-layer}
    \label{fig:1hl-clas-acc}
  \end{subfigure}
  \hfill
  \vspace{-0.4cm}

  \begin{subfigure}[b]{1.0\linewidth}
    \centering
    \begin{overpic}[width=\linewidth,height=3.5cm]{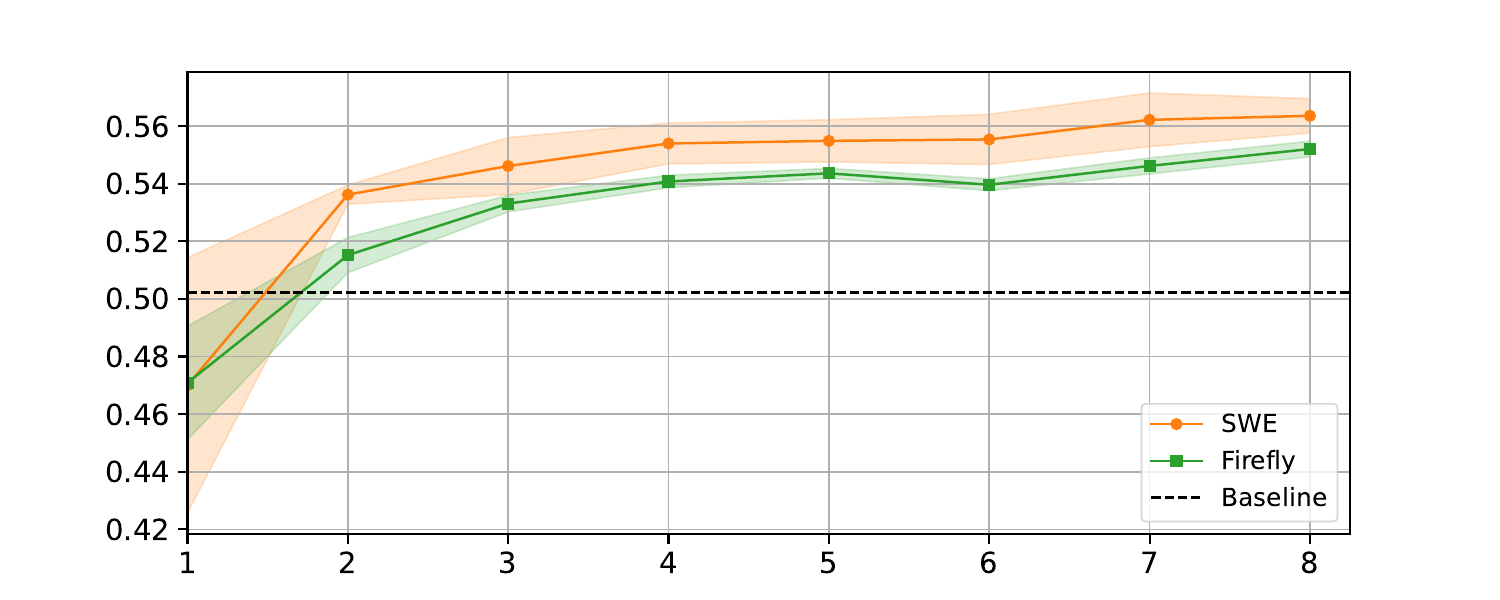}
      \put(3,16){\rotatebox{90}{\text{\scriptsize Accuracy}}}
      \put(50,0){\makebox(0,0){\text{\scriptsize Stages}}}
    \end{overpic}
    \caption{Three hidden-layers}
    \label{fig:3hl-clas-acc}
  \end{subfigure}
  
  \caption{Test accuracy after each expansion stage on CIFAR-100. 
  SWE and SWE+SVoD consistently outperform Firefly in the single hidden-layer setting and surpass the fixed-size baseline in early stages. 
  Shaded regions indicate standard deviation.}
  \label{fig:cifar-100-cla}
\end{figure}

We evaluate classification performance under the same incremental growth and training protocol as in Section~\ref{sec:imagereconstruction}. First, we consider a single hidden-layer network Table \ref{tab:classification_single}. We then extend the evaluation to a deeper architecture with three hidden layers, each initialized with 10 neurons. In this case, neuron allocation during expansion is guided by the proposed Steepest Voting Distributor (SVoD). To assess its effectiveness, we also compare against a random allocation strategy denoted as RAS in Table \ref{tab:classification_three}, where new neurons are distributed uniformly across layers, highlighting the superior performance of SVoD in directing neuron placement. In Figures \ref{fig:1hl-clas-acc}, \ref{fig:3hl-clas-acc} we demonstrate more analytically the evolution of test accuracy during successive expansion stages. 



\begin{table}[htb!]
\centering

\begin{subtable}{\linewidth}
\centering
\small
\begin{tabular}{@{}lccc@{}}
\toprule
Dataset & Baseline & Firefly & SWE  \\
\midrule
MNIST        & $97.54 \pm 0.17$ & $96.80 \pm 0.23$ & $\mathbf{97.62 \pm 0.12}$  \\
FMNIST  & $\mathbf{85.78 \pm 0.12}$ & $83.79 \pm 0.28$ & $84.81 \pm 0.51$           \\
CIFAR10     & $82.06 \pm 0.17$  & $82.68 \pm 0.43$ & $\mathbf{83.72 \pm 0.13}$  \\
CIFAR100     & $53.87 \pm 0.42$  & $55.78 \pm 0.57$ & $\mathbf{59.57  \pm 0.23}$ \\
\bottomrule
\end{tabular}
\caption{MLPs with single hidden layer}
\label{tab:classification_single}
\end{subtable}

\begin{subtable}{\linewidth}
\vspace{0.3cm}
\centering
\small
\resizebox{\linewidth}{!}{
\begin{tabular}{@{}lcccc@{}}
\toprule
Dataset & Baseline & Firefly & SWE+RAS & SWE+SVoD  \\
\midrule
MNIST     & $97.97 \pm 0.23$    & $97.88 \pm 0.17$ & $97.93\pm0.25$ &  $\mathbf{98.10 \pm 0.27}$  \\
FMNIST & $\mathbf{86.35 \pm 0.15}$  & $85.45 \pm 0.11$ & $85.36\pm0.49$ & $85.66 \pm 0.29$           \\
CIFAR10    & $82.20 \pm 0.27$   & $83.21 \pm 0.60$ & $82.46\pm0.36$ & $\mathbf{83.32 \pm 0.39}$ \\
CIFAR100  & $49.60 \pm 0.57$    & $55.20 \pm 0.27$ & $56.11\pm0.25$ & $\mathbf{ 56.36 \pm 0.60}$  \\
\bottomrule
\end{tabular}
}
\caption{MLPs with three hidden layers}
\label{tab:classification_three}
\end{subtable}
\caption{Final test accuracy (\%) after all growth stages for image classification. Our methods, SWE and SWE+SVoD, generally outperform both fixed-size baselines and the Firefly method. For CIFAR datasets, MLP heads are applied on a backbone CNN that is trained during the procedure but has fixed size.}
\label{tab:classification_results}
\end{table}




For CIFAR-10 and CIFAR-100, we adopt a lightweight VGG-style CNN backbone consisting of two convolutional blocks. Each block has two $3\times3$ convolutional layers with batch normalization and ReLU activation, followed by $2\times2$ max pooling. In CIFAR-10, the blocks increase channels from $3 \rightarrow 16$ and $16 \rightarrow 32$, while in CIFAR-100 they increase from $3 \rightarrow 32$ and $32 \rightarrow 64$. The output of the backbone is flattened and passed to a fully connected MLP head, which is configured with either one or three hidden layers as described above. For CIFAR-100, each hidden layer starts with four times as many neurons as in the other datasets. The extension mechanism is applied only to the MLP head.

We observe that our method consistently outperforms expansion strategies that lack a smooth adjustment phase. In most cases, it also surpasses fixed-size baselines, demonstrating the effectiveness of augmenting capacity through the smooth integration of new neurons. Furthermore, the proposed distributing scheme outperforms agnostic layer allocation (SWE+RAS), and its random-search-based voting mechanism (SWE+SVoD) makes our overall framework a competitive alternative to prior expansion methods.





\section{CONCLUSION}
\label{sec:conclusion}
In this work we addressed the challenge of progressive network growth, where the effective use of newly added neurons is often hindered, in part due to their tendency to remain inactive. Across the experimental settings we considered, our methods substantially reduced this problem and delivered strong results. They consistently outperformed other competitive expansion strategies, and in some cases even surpassed fixed-size baselines trained with the standard approach.

In future work, SWE could be applied to continual learning scenarios, where expanding network capacity before training on new tasks may help reduce interference and forgetting. Second, the same mechanism that SWE uses to integrate new neurons could be adapted for pruning: by introducing a regularization term that encourages neurons to redistribute their weights to others, we could enforce smooth disappearance of those neurons.

\section{Compliance with Ethical Standards}
\label{sec:Compliance with Ethical Standards}

This is a numerical simulation study for which no ethical approval was required.


\bibliographystyle{IEEEbib}
\bibliography{refs}

\begin{thebibliography}{10}

\bibitem{zoph2017neural}
Barret Zoph and Quoc Le,
\newblock ``Neural architecture search with reinforcement learning,''
\newblock in {\em International Conference on Learning Representations}, 2017.

\bibitem{pham2018enas}
Hieu Pham, Melody Guan, Barret Zoph, Quoc Le, and Jeff Dean,
\newblock ``Efficient neural architecture search via parameters sharing,''
\newblock in {\em Proceedings of the 35th International Conference on Machine Learning (ICML)}, 2018.

\bibitem{liu2018darts}
Hanxiao Liu, Karen Simonyan, and Yiming Yang,
\newblock ``{DARTS}: Differentiable architecture search,''
\newblock in {\em International Conference on Learning Representations}, 2019.

\bibitem{10.1007/978-3-030-58555-6_28}
Xiangxiang Chu, Tianbao Zhou, Bo~Zhang, and Jixiang Li,
\newblock ``Fair darts: Eliminating unfair advantages in differentiable architecture search,''
\newblock in {\em Computer Vision – ECCV 2020: 16th European Conference, Glasgow, UK, August 23–28, 2020, Proceedings, Part XV}, Berlin, Heidelberg, 2020, p. 465–480, Springer-Verlag.

\bibitem{Han2015LearningBW}
Song Han, Jeff Pool, John Tran, and William~J. Dally,
\newblock ``Learning both weights and connections for efficient neural network,''
\newblock in {\em Neural Information Processing Systems}, 2015.

\bibitem{NEURIPS2020_fdbe012e}
Lemeng Wu, Bo~Liu, Peter Stone, and Qiang Liu,
\newblock ``Firefly neural architecture descent: a general approach for growing neural networks,''
\newblock in {\em Advances in Neural Information Processing Systems}, H.~Larochelle, M.~Ranzato, R.~Hadsell, M.F. Balcan, and H.~Lin, Eds. 2020, vol.~33, pp. 22373--22383, Curran Associates, Inc.

\bibitem{evci2022gradmax}
Utku Evci, Bart van Merrienboer, Thomas Unterthiner, Fabian Pedregosa, and Max Vladymyrov,
\newblock ``Gradmax: Growing neural networks using gradient information,''
\newblock in {\em International Conference on Learning Representations}, 2022.

\bibitem{Rusu2016ProgressiveNN}
Andrei~A. Rusu, Neil~C. Rabinowitz, Guillaume Desjardins, Hubert Soyer, James Kirkpatrick, Koray Kavukcuoglu, Razvan Pascanu, and Raia Hadsell,
\newblock ``Progressive neural networks,''
\newblock {\em arXiv preprint arXiv:1606.04671}, 2016.

\bibitem{yoon2018lifelong}
Jaehong Yoon, Eunho Yang, Jeongtae Lee, and Sung~Ju Hwang,
\newblock ``Lifelong learning with dynamically expandable networks,''
\newblock in {\em International Conference on Learning Representations}, 2018.

\bibitem{Liu2019SplittingSD}
Qiang Liu, Lemeng Wu, and Dilin Wang,
\newblock ``Splitting steepest descent for growing neural architectures,''
\newblock in {\em Neural Information Processing Systems}, 2019.

\bibitem{wang2020energyawareneuralarchitectureoptimization}
Dilin Wang, Meng Li, Lemeng Wu, et~al.,
\newblock ``Energy-aware neural architecture optimization with fast splitting steepest descent,'' 2020.

\bibitem{mitchell2024selfexpandingneuralnetworks}
Rupert Mitchell, Robin Menzenbach, Kristian Kersting, and Martin Mundt,
\newblock ``Self‐expanding neural networks,''
\newblock {\em arXiv preprint arXiv:2307.04526}, 2023.

\bibitem{Chen2015Net2NetAL}
Tianqi Chen, Ian~J. Goodfellow, and Jonathon Shlens,
\newblock ``Net2net: Accelerating learning via knowledge transfer,''
\newblock {\em CoRR}, vol. abs/1511.05641, 2015.

\bibitem{cimb44020056}
Miao Wang, Xu~Yang, Yunchong Qian, et~al.,
\newblock ``Adaptive neural network structure optimization algorithm based on dynamic nodes,''
\newblock {\em Current Issues in Molecular Biology}, vol. 44, no. 2, pp. 817--832, 2022.

\bibitem{deadrelu}
Lu~Lu, Yeonjong Shin, Yanhui Su, and George Karniadakis,
\newblock ``Dying relu and initialization: Theory and numerical examples,''
\newblock {\em Communications in Computational Physics}, vol. 28, pp. 1671--1706, 11 2020.

\bibitem{6296535}
Li~Deng,
\newblock ``The mnist database of handwritten digit images for machine learning research [best of the web],''
\newblock {\em IEEE Signal Processing Magazine}, vol. 29, no. 6, pp. 141--142, 2012.

\bibitem{xiao2017fashionmnistnovelimagedataset}
Han Xiao, Kashif Rasul, and Roland Vollgraf,
\newblock ``Fashion-mnist: a novel image dataset for benchmarking machine learning algorithms,''
\newblock {\em arXiv preprint arXiv:1708.07747}, 2017.

\bibitem{cifar10}
Alex Krizhevsky,
\newblock ``Learning multiple layers of features from tiny images,''
\newblock Tech. {R}ep., University of Toronto, 2009.

\bibitem{Krizhevsky2009LearningML}
Alex Krizhevsky,
\newblock ``Learning multiple layers of features from tiny images,''
\newblock 2009.

\bibitem{adam_optimizer}
Diederik~P Kingma and Jimmy Ba,
\newblock ``Adam: A method for stochastic optimization,''
\newblock {\em arXiv preprint arXiv:1412.6980}, 2014.

\bibitem{He2015DelvingDI}
Kaiming He, X.~Zhang, Shaoqing Ren, and Jian Sun,
\newblock ``Delving deep into rectifiers: Surpassing human-level performance on imagenet classification,''
\newblock {\em 2015 IEEE International Conference on Computer Vision (ICCV)}, pp. 1026--1034, 2015.

\end{thebibliography}

\vspace{0.1cm}

\begin{center}
    \small\textbf{ACKNOWLEDGMENT}
\end{center}
\begin{minipage}{0.7\linewidth}
\small
We thank Dr. Panagiotis Filntisis, Postdoctoral Associate at Robotics Institute/Athena RC and HERON - Hellenic Robotics Center of Excellence, for insightful feedback and constructive suggestions that greatly improved this work.

This project is partially funded by the European Union under Horizon Europe (grant No. 101136568 -- HERON).
\end{minipage}%
\hfill
\begin{minipage}{0.20\linewidth}
    \includegraphics[width=\linewidth]{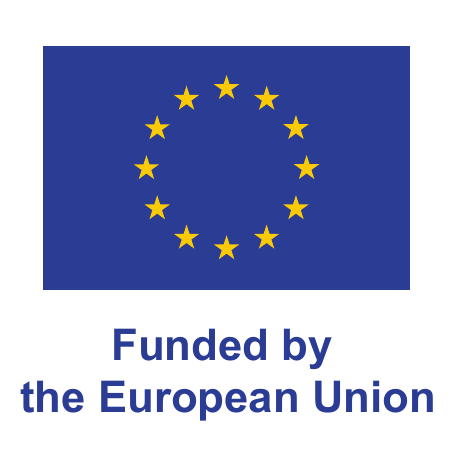}
\end{minipage}

\end{document}